\documentclass{article} 
\usepackage{iclr2021_conference,times}

\usepackage{booktabs}
\usepackage{multirow}
\usepackage{tabularx}
\usepackage{xcolor}
\usepackage{caption} 
\usepackage{tikz}
\usetikzlibrary{shapes.geometric, arrows}

\tikzstyle{startstop} = [rectangle, rounded corners, minimum width=2.5cm, minimum height=1cm,text centered, draw=black, fill=red!30]
\tikzstyle{process} = [rectangle, minimum width=2.5cm, minimum height=1cm, text centered, draw=black, fill=orange!30]
\tikzstyle{decision} = [diamond, minimum width=2.5cm, minimum height=1cm, text centered, draw=black, fill=green!30]
\tikzstyle{arrow} = [thick,->,>=stealth]

\usepackage{amsmath,amsfonts,bm}

\def\eqref#1{equation~\ref{#1}}

\def\1{\bm{1}}

\DeclareMathAlphabet{\mathsfit}{\encodingdefault}{\sfdefault}{m}{sl}
\SetMathAlphabet{\mathsfit}{bold}{\encodingdefault}{\sfdefault}{bx}{n}

\usepackage{hyperref}
\usepackage{url}
\usepackage{algorithm}
\usepackage{algpseudocode}
\usepackage{authblk}  

\title{Multi-Programming Language Ensemble for Code Generation in Large Language Model}

\author{Tengfei Xue}  

\author{Xuefeng Li}  
  
\author{Tahir Azim}  
  
\author{Roman Smirnov}  
  
\author{Jianhui Yu}  
  
\author{\authorcr Arash Sadrieh}  
  
\author{Babak Pahlavan}  
\affil{NinjaTech AI} 

\iclrfinalcopy 
\begin{document}

\maketitle
\begin{abstract}

Large language models (LLMs) have significantly improved code generation, particularly in one-pass code generation. However, most existing approaches focus solely on generating code in a single programming language, overlooking the potential of leveraging the multi-language capabilities of LLMs. LLMs have varying patterns of errors across different languages, suggesting that a more robust approach could be developed by leveraging these multi-language outputs. In this study, we propose Multi-Programming Language Ensemble (MPLE), a novel ensemble-based method that utilizes code generation across multiple programming languages to enhance overall performance. By treating each language-specific code generation process as an individual ``weak expert" and effectively integrating their outputs, our method mitigates language-specific errors and biases. This multi-language ensemble strategy leverages the complementary strengths of different programming languages, enabling the model to produce more accurate and robust code. Our approach can be seamlessly integrated with commonly used techniques such as the reflection algorithm and Monte Carlo tree search to improve code generation quality further. Experimental results show that our framework consistently enhances baseline performance by up to 17.92\% on existing benchmarks (HumanEval and HumanEval-plus), with a standout result of 96.25\% accuracy on the HumanEval benchmark, achieving new state-of-the-art results across various LLM models. The code will be released at \href{https://github.com/NinjaTech-AI/MPLE}{https://github.com/NinjaTech-AI/MPLE}

\end{abstract}

\section{Introduction}
\label{sec_intro}

Large Language Models (LLMs) have significantly advanced the field of code generation, demonstrating impressive capabilities in generating syntactically correct and semantically meaningful code across various programming languages~\citep{chen2021evaluating, li2022competition, austin2021program, liu2024your}. Recent progress has been marked by the ability of these models, such as GPT 4~\citep{achiam2023gpt}, Llama 3~\citep{dubey2024llama}, and Claude 3~\citep{anth2024claude}, to produce high-quality code snippets from natural language descriptions, often excelling in specific languages like Python or Java~\citep{li2023starcoder, roziere2023code, zhong2024debug, huang2023agentcoder, islam2024mapcoder}. However, the majority of existing approaches in code generation have primarily focused on a single programming language, neglecting the potential advantages of leveraging multi-language capabilities to enhance the robustness and accuracy of generated code.

LLMs exhibit varying error patterns across different programming languages due to differences in syntax, semantics, and idiomatic practices~\citep{peng2024humaneval, zheng2023codegeex, athiwaratkunmulti, cassano2022multipl}. For example, an LLM may perform well in Python code generation but generate errors in Java or C++ due to differences in error handling or library usage. These variations indicate that LLMs have language-specific biases, which could be mitigated through a more robust, multi-language approach. By leveraging outputs generated across different programming languages, it is possible to reduce these biases and improve the overall performance of code generation.

In this study, we introduce Multi-Programming Language Ensemble (MPLE), a novel ensemble-based method for code generation that harnesses the multi-language capabilities of LLMs. Inspired by ensemble learning techniques in machine learning, where multiple models are combined to form a stronger, more accurate model, we treat each language-specific code generation task as a ``weak expert" and utilize the outputs from multiple languages to iteratively improve the overall performance. By effectively integrating the outputs from these different experts, our method aims to mitigate language-specific errors and biases, thereby enhancing the robustness and accuracy of the generated code.

Our framework integrates a programming language sampling algorithm to guide the code generation process. Starting with an initial code generation in a chosen programming language, the model is prompted to produce alternative versions in other languages when errors are detected. These alternative versions are translated back to the original language to exploit complementary strengths and mitigate language-specific weaknesses. This iterative process continues until all visible/internal tests are passed or a maximum number of language transformations is reached, ensuring a thorough exploration of potential solutions.

Furthermore, we demonstrate how to seamlessly integrate our ensemble strategy with existing techniques such as the reflection algorithm~\citep{shinn2024reflexion, yao2023react} and Monte Carlo Tree Search (MCTS)~\citep{chaslot2008monte, zhoulanguage, zhang2024accessing}, which improve reasoning and decision-making capabilities in LLMs. By integrating these methods, we aim to enhance the quality of code generation further and expand the capabilities of LLMs in handling complex programming tasks.

Our contributions in this paper are threefold: (1) We introduce a multi-language ensemble framework for code generation in LLMs, leveraging the strengths of different programming languages to improve robustness and accuracy; (2) We demonstrate how this framework can be integrated with existing methods such as the reflection algorithm and MCTS to further enhance code quality; (3) We validate our approach through extensive experiments on benchmarks such as HumanEval and HumanEval-plus datasets, achieving new state-of-the-art results and improvements by up to 17.92\% across various LLM models.

\begin{figure}[t]
\centering
\includegraphics[width=\textwidth]{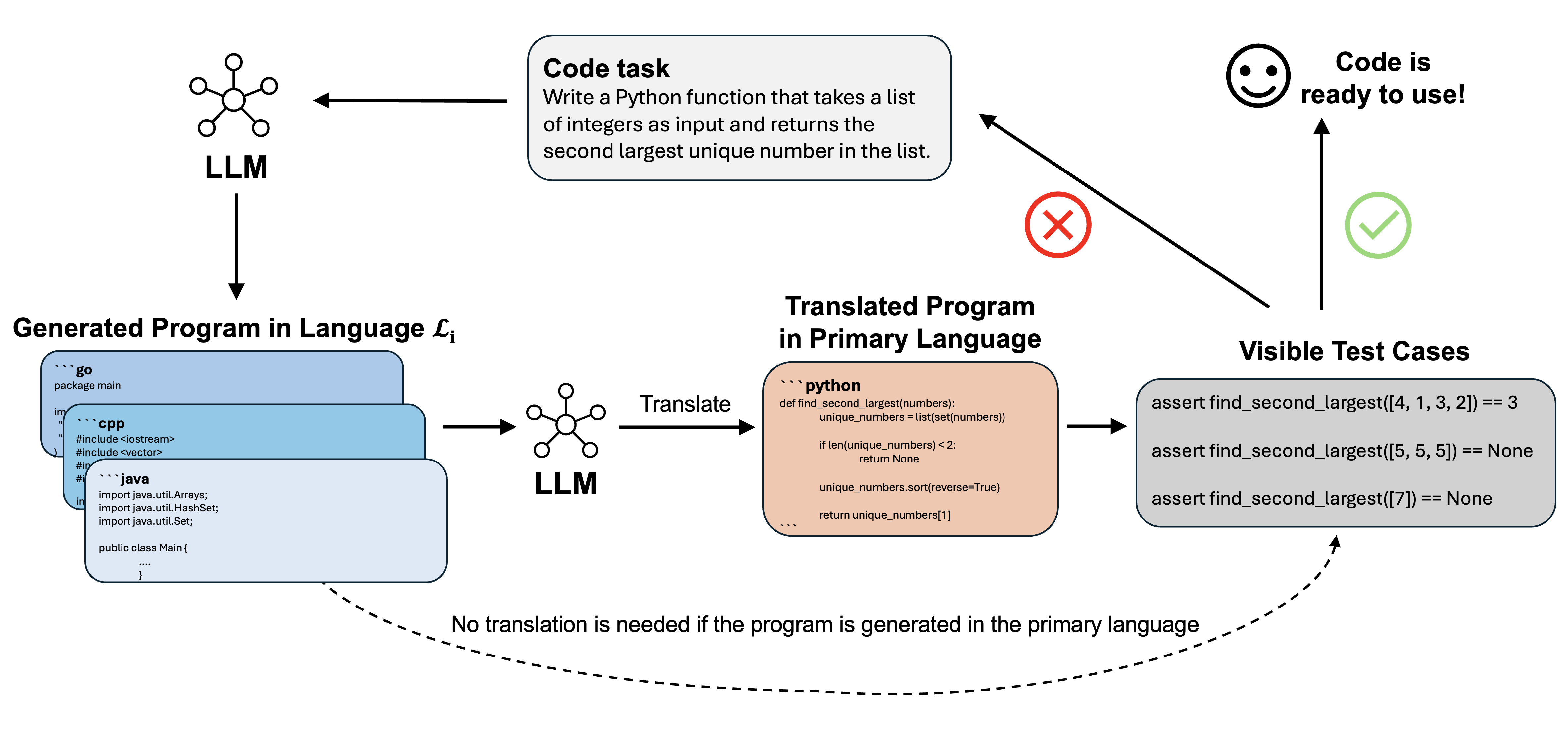} 
\caption{Overview of the Multi-Programming Language Ensemble (MPLE) framework for code generation.}
\label{fig_overview}
\end{figure}

\section{Methodology}
\label{sec_method}

In this section, we present our Multi-Programming Language Ensemble (MPLE) framework (Fig.~\ref{fig_overview}) for code generation in Large Language Models (LLMs). This approach iteratively refines code by leveraging the strengths of different programming languages, reducing language-specific errors and biases. We integrate this method with reflection algorithms and MCTS to enhance the overall robustness and accuracy of the generated code. The following subsections provide a detailed description of the methodology.

\subsection{Problem Formulation}

We follow the problem formulation used in~\cite{zhong2024debug}. We formulate the code generation task as follows: Each sample can be represented as a triplet $(Q, T_v, T_h)$, where $Q$ is the code task description, $T_v$ represents the visible test cases, and $T_h$ denotes the hidden test cases. At the outset, the LLM is provided with $Q$ and $T_v$ to generate an initial program, $P_0$. The generated program $P_0$ is then refined iteratively to produce a sequence of programs $\{P_1, P_2, \ldots, P_n\}$ until a program passes all the visible tests in $T_v$. The final output program is denoted as $P^*$. This final program, $P^*$, is then evaluated on the hidden test cases $T_h$ to verify its correctness. Notably, $T_h$ is used only once (pass@1) and remains hidden during the code generation and refinement process.

\subsection{Framework Overview}

The proposed MPLE framework (Fig.~\ref{fig_overview}) is designed to utilize the multi-language capabilities of LLMs to improve code generation. The process consists of several steps:

\begin{enumerate}
    \item \textbf{Initial Code Generation:} The process begins by prompting the LLM to generate an initial code version $P_0$ in a primary programming language $\mathcal{L}_0$ based on the given task description $Q$. This generated code $P_0$ is then tested against the visible test cases $T_v$. If $P_0$ passes all visible tests, the code generation process is terminated, and $P_0$ is further evaluated on the hidden tests $T_h$ to determine the final result.
    
    \item \textbf{Multi-Language Sampling and Translation:} If $P_0$ fails to pass all visible test cases, the framework prompts the LLM to generate a new code version $P_{L_i}$ in a different programming language $\mathcal{L}_i$ (e.g., if $P_0$ is in Python, $P_{L_i}$ could be generated in Java). The generated code $P_{L_i}$ is then translated back into the original programming language $\mathcal{L}_0$ to produce a refined version $P_i$. This refined version is designed to maintain the logical structure of the newly generated code while conforming to the syntax and semantics of the primary language.
    
    \item \textbf{Iterative Refinement:} The refined code version $P_i$ is tested against the visible test cases $T_v$. If it passes all tests, the process is terminated, and $P_i$ is evaluated on the hidden tests $T_h$. If $P_i$ fails to pass all visible tests, the framework continues by generating an additional code version $P_{L_{i+1}}$ in another programming language (e.g., C++). The new version $P_{L_{i+1}}$ is then translated back into the primary language to produce a further refined version $P_{i+1}$. This iterative process continues, utilizing different programming languages, until a code version passes all visible tests or the maximum number of languages ($L_{\text{max}}$) is reached.
    
    \item \textbf{Ensemble Integration:} Throughout the iterations, the ensemble framework integrates the strengths of multiple languages to progressively refine the program. By treating each language-specific code generation as an individual ``weak expert," the framework combines their outputs to mitigate language-specific errors and biases. This approach leverages the unique strengths of different programming languages, such as differences in syntax, semantics, and idiomatic usage, to produce more robust and accurate code. If no version passes all visible tests within $L_{\text{max}}$, the last generated version $P_{L_{\text{max}}}$ is evaluated on the hidden tests $T_h$ to determine the final result.
\end{enumerate}

The overall process of our MPLE framework can be summarized in Algorithm~\ref{alg:multi-lang-ensemble}.

\begin{algorithm}
\caption{MPLE for Code Generation}
\label{alg:multi-lang-ensemble}
\begin{algorithmic}[1]
\Require $Q, T_v, T_h, \mathcal{L}, L_{\text{max}}$ \Comment{Task description, visible and hidden tests, set of languages, max number of languages}
\Ensure Result indicating \textit{succeed} or \textit{fail}

\For{$i \gets 0$ \textbf{to} $L_{\text{max}}$}
    \State $P_i \gets \text{LLM}(Q, \mathcal{L}_i)$ \Comment{Generate program $P_i$ in language $\mathcal{L}_i$}
    \If{$\mathcal{L}_i$ \textbf{is not} primary language}
        \State $P_i \gets \text{translate}(P_i)$ \Comment{Translate $P_i$ back to primary language}
    \EndIf
    \If{$\text{eval}(P_i, T_v) = 1$}
        \If{$\text{eval}(P_i, T_h) = 1$}
            \State \Return \textit{succeed} \Comment{Passes both visible and hidden tests}
        \Else
            \State \Return \textit{fail} \Comment{Fails hidden tests}
        \EndIf
    \EndIf
\EndFor

\State \Return \textit{fail} \Comment{All attempts failed}
\end{algorithmic}
\end{algorithm}

\subsection{Integration with Existing Techniques}

To further enhance the code generation process, our ensemble framework seamlessly integrates with existing techniques such as the reflection algorithm~\citep{shinn2024reflexion} and Monte Carlo Tree Search (MCTS)~\citep{chaslot2008monte}. These integrations allow for a more dynamic and iterative refinement of the generated code, ultimately improving the robustness and accuracy of the results.

\begin{itemize}
    \item \textbf{Reflection Algorithm:} The reflection algorithm uses feedback from the execution of visible test cases to iteratively refine the code. Our MPLE framework is integrated into the reflection algorithm by utilizing its iterative refinement process. In each iteration, MPLE generates a code version using multiple programming languages. The code is tested against the visible test cases, and any failures or errors are used as feedback to prompt further refinements. This process of reflection allows the model to continuously learn from its mistakes, reducing language-specific errors and enhancing the quality of the generated code across multiple iterations. The overall process of integrating our MPLE framework with reflection algorithm is in Appendix~\ref{app_reflection_algo}.
    
    \item \textbf{MCTS:} MCTS builds a decision tree where every node in the tree is a state and the edge is an action. MCTS is applied to explore different code generation paths together with our MPLE framework. The integration of MPLE with MCTS involves representing each code version generated by the MPLE framework as a node in the MCTS search tree. MCTS systematically explores different code generation paths by selecting, expanding, and simulating nodes that correspond to different code versions. This integration helps efficiently search for the most promising code paths, leveraging both the exploration capabilities of MCTS and the language-ensemble ability of MPLE. The overall process of integrating our MPLE framework with MCTS is in Appendix~\ref{app_mcts_algo}

\end{itemize}

By combining these techniques, we enhance the ensemble framework's ability to generate accurate and robust code, leveraging both iterative improvement and strategic exploration.

\section{Experiments}
\label{sec_experiments}

We evaluate our proposed MPLE framework on two widely recognized code generation benchmarks: HumanEval~\citep{chen2021evaluating} and HumanEval-plus~\citep{liu2024your}. These benchmarks assess the capability of large language models (LLMs) to generate functional code based on textual descriptions.

HumanEval is designed for text-to-code (Python) generation tasks where the input is a brief passage describing the intended functionality of the program to be generated. The output is then evaluated based on its ability to pass unit tests with specified requirements. HumanEval-plus extends the HumanEval dataset by incorporating a large number of additional valid unit test cases to rigorously evaluate the synthesized code's robustness and correctness.

\subsection{Experimental Setup}

We compute Pass@1 accuracy using hidden test cases to assess the performance of the generated code. Pass@1 measures the percentage of tasks for which the model's top output passes all hidden test cases, providing a stringent evaluation metric for the models' capability to generate correct code.

We conducted experiments using both proprietary and open-source LLMs:

\begin{itemize}
    \item \textbf{Proprietary LLMs}: GPT3.5-turbo (gpt-3.5-turbo-0125), GPT-4o-mini (gpt-4o-mini-2024-07-18), GPT-4o (gpt-4o-2024-05-13), and Claude-Sonnet-3.5.
    \item \textbf{Open-source LLMs}: Llama3.1-8b-instruct, Llama3.1-70b-instruct, and Llama3.1-405b-instruct.
\end{itemize}

\subsection{Methods Evaluated}

We evaluated the performance of the following methods:

\begin{enumerate}
    \item \textbf{Baseline}: The model is directly prompted to generate code based on the task description without additional strategies. This serves as a benchmark for comparing the effectiveness of more sophisticated approaches.
   
    \item \textbf{MPLE}: Our proposed method integrates Java and C++ into Python programming, allowing the model to utilize multi-language capabilities for code generation. Java and C++ are selected here because LLMs generally have high performance in these programming languages~\citep{peng2024humaneval, zheng2023codegeex, athiwaratkunmulti, cassano2022multipl}. Note that MPLE is able to integrate with any number of programming languages. The ensemble approach aims to improve code accuracy by leveraging the strengths of multiple programming languages.

    \item \textbf{MPLE+Reflection}: This method combines the proposed MPLE strategy with the reflection algorithm~\citep{shinn2024reflexion}, enabling iterative self-correction and refinement. The maximum number of iterations is set to 8, providing the model with multiple opportunities to refine its output based on feedback from visible test cases.

    \item \textbf{MPLE+MCTS}: This method integrates the proposed MPLE strategy with MCTS~\citep{chaslot2008monte} to explore the search space of possible code solutions more effectively. The MCTS algorithm runs with a maximum of 8 iterations and 5 nodes each iteration, allowing the model to systematically explore different code generation paths and select the most promising ones.
\end{enumerate}

\subsection{Results}

The performance results of each method on the HumanEval and HumanEval-plus benchmarks are presented in Tables~\ref{table:humaneval} and~\ref{table:humanevalplus}, respectively. The results demonstrate the impact of each method on Pass@1 accuracy.

\begin{table}[htbp]
\centering
\caption{Performance comparison on HumanEval benchmark for various LLMs using different methods. Best results for this benchmark are in \textbf{bold}.}
\label{table:humaneval}
\begin{tabular}{lcccc}
\hline
\textbf{Model} & \textbf{Baseline} & \textbf{MPLE} & \textbf{MPLE+Reflection} & \textbf{MPLE+MCTS} \\
\hline
GPT3.5-turbo        & 65.83\%           & 74.17\%  & 80.00\%      & 83.75\% \\
GPT-4o-mini         & 87.71\%           & 88.75\%  & 91.87\%      & 93.12\% \\
GPT-4o              & 90.62\%           & 91.67\%  & 94.37\%      & 95.00\% \\
Claude-Sonnet-3.5   & 86.88\%         & 88.75\%    & 93.13\%      & 93.13\% \\
llama3.1-8b-instruct & 66.87\%         & 71.88\%   & 77.50\%      & 75.00\% \\
llama3.1-70b-instruct & 78.80\%        & 85.21\%   & 89.38\%      & 92.50\% \\
llama3.1-405b-instruct & 86.46\%       & 93.44\%  & 95.63\%      & \textbf{96.25\%} \\
\hline
\end{tabular}
\end{table}

\begin{table}[htbp]
\centering
\caption{Performance comparison on HumanEval-plus benchmark for various LLMs using different methods. Best results for this benchmark are in \textbf{bold}.}
\label{table:humanevalplus}
\begin{tabular}{lcccc}
\hline
\textbf{Model} & \textbf{Baseline} & \textbf{MPLE} & \textbf{MPLE+Reflection} & \textbf{MPLE+MCTS} \\
\hline
GPT3.5-turbo        & 61.04\%      & 61.88\%        & 73.75\%        & 71.88\% \\
GPT-4o-mini         & 81.87\%      & 82.50\%        & \textbf{87.50\%}        & 86.67\%   \\
GPT-4o              & 83.75\%      & 85.21\%        & 84.38\%        & \textbf{87.50\%}   \\
Claude-Sonnet-3.5   & 82.50\%      & 86.25\%        & 86.88\%        & \textbf{87.50\%}   \\
llama3.1-8b-instruct & 60.00\%   & 66.25\% & 71.88\%        & 68.75\%   \\
llama3.1-70b-instruct & 78.75\%          & 82.50\%        & 85.00\%        & 83.75\%         \\
llama3.1-405b-instruct & 80.63\% & 86.25\% & \textbf{87.50\%}        & \textbf{87.50\%}  \\
\hline
\end{tabular}
\end{table}

Table~\ref{table:humaneval} shows the performance on the HumanEval benchmark. Our proposed MPLE framework consistently improves the Pass@1 accuracy across all tested LLMs compared to the Baseline approach. For example, GPT3.5-turbo's accuracy increased from 65.83\% in the Baseline to 74.17\% with MPLE, highlighting the effectiveness of leveraging multiple programming languages to reduce language-specific biases and errors. Furthermore, integrating MPLE with advanced inference techniques like Reflection and MCTS yields additional performance gains. The combination of MPLE+MCTS achieved the highest accuracy for several models, such as llama3.1-405b-instruct, which reached a SOTA Pass@1 accuracy of 96.25\% on this benchmark. These results indicate that MPLE, especially when combined with other inference algorithms, provides a robust framework for enhancing code generation in LLMs.

Table~\ref{table:humanevalplus} provides the performance on the HumanEval-plus benchmark, further validating the benefits of our multi-language ensemble approach. Similar to the HumanEval results, MPLE demonstrates consistent improvements over the Baseline across all tested models. Notably, llama3.1-8b-instruct's performance improved from 60.00\% in the Baseline to 71.88\% with MPLE+Reflection, showing the strength of combining MPLE with reflection-based iterative refinement. Additionally, MPLE+Reflection and MPLE+MCTS deliver competitive results, with multiple models (GPT-4o-mini, GPT-4o, Claude-Sonnet-3.5, and llama3.1-405b-instruct) achieving 87.50\%. 

The experimental results suggest that the MPLE framework, especially when used in conjunction with additional inference algorithms, offers a powerful and flexible approach for enhancing code generation across various LLMs. This approach's consistent performance improvements and state-of-the-art achievements underscore its potential for practical applications in AI-driven software development.


\section{Conclusion}

In this paper, we propose MPLE, a novel multi-programming language ensemble framework for code generation in Large Language Models (LLMs). Our approach leverages the strengths of multiple programming languages to iteratively refine code generation, thereby enhancing the overall performance and robustness of the models. By integrating strategies such as the reflection algorithm and MCTS with our ensemble framework, we demonstrate significant improvements in Pass@1 accuracy across multiple benchmarks, including HumanEval and HumanEval-plus. The experimental results demonstrate that our method consistently outperforms baseline models, which effectively explores optimal code solutions. Our MPLE approach reduces language-specific errors and harnesses the unique strengths of various programming languages, resulting in more accurate and robust code generation. These findings suggest that combining multi-language ensembles with iterative refinement is a promising direction for advancing code generation in LLMs. Our framework can be further developed to address more complex coding tasks and diverse programming environments, contributing to the evolution of AI-driven software development.

Future work will focus on integrating more efficient token generation strategies~\citep{xue2024ninjallm, kim2024speculative} and more advanced inference algorithms~\citep{wang2024litesearch} to further enhance code generation. We also plan to evaluate our approach on a broader range of datasets and real-world challenges~\citep{tian2024debugbench, jimenezswe} to assess its generalizability. Additionally, we will explore how to effectively deploy our framework in a production environment, ensuring it meets practical performance and reliability requirements. Currently, our NinjaLLM 3.0 (a fine-tuned and quantized version of llama3.1-405b) has achieved promising scores on HumanEval (93.85\%) and HumanEval-plus (86.67\%), and we are on the path to further improving its performance.

\newpage

\bibliography{iclr2021_conference}
\bibliographystyle{iclr2021_conference}

\newpage

\appendix
\section{Appendix}
\label{app_reflection_algo}
\begin{algorithm}
\caption{Reflection Algorithm with MPLE Integration}
\begin{algorithmic}[1]
\Require $Q$, $T_v$, $T_h$, $max\_iterations$~~~~\Comment{Task description $Q$, visible test cases $T_v$, hidden test cases $T_h$, max iterations}
\Ensure Result indicating \textit{succeed} or \textit{fail}

\State $i \gets 0$ \Comment{Initialize iteration counter}
\State $max\_iterations \gets k$ \Comment{Set maximum iterations}

\While{$i < max\_iterations$}
    \State $P_i \gets \text{MPLE}(Q)$ \Comment{Generate program $P_i$ using MPLE}
    \If{$\text{eval}(P_i, T_v) = 1$}
        \If{$\text{eval}(P_i, T_h) = 1$}
            \State \Return \textit{succeed} \Comment{Return ``succeed" if all tests pass}
        \Else
            \State \Return \textit{fail} \Comment{Return ``fail" if hidden tests fail}
        \EndIf
    \Else
        \State $feedback \gets \text{get\_feedback}(T_v, result)$ \Comment{Extract feedback from failed visible tests}
        \State $P_{i+1} \gets \text{MPLE}(Q, feedback)$ \Comment{Refine $P_i$ with MPLE based on feedback}
    \EndIf
    \State $i \gets i + 1$ \Comment{Increment iteration}
\EndWhile

\State \Return \textit{fail} \Comment{Return ``fail" if maximum iterations reached without passing all tests}
\end{algorithmic}
\end{algorithm}

\section{Appendix}
\label{app_mcts_algo}
\begin{algorithm}
\caption{MCTS Integration with MPLE}
\begin{algorithmic}[1]
\Require $Q$, $T_v$, $T_h$, $max\_iterations$, $node\_expansion$ \Comment{Task description $Q$, visible test cases $T_v$, hidden test cases $T_h$, max iterations, node expansion factor}
\Ensure Result indicating \textit{succeed} or \textit{fail}

\State Initialize tree with root node $n_0$ representing initial program $P_0$
\State $i \gets 0$ \Comment{Initialize iteration counter}

\While{$i < max\_iterations$}
    \State $node \gets \text{select\_node}(tree)$ \Comment{Select a node to expand based on the tree policy}
    
    \State $P_i \gets \text{MPLE}(Q)$ \Comment{Generate program $P_i$ using MPLE at the selected node}
    
    \If{$\text{eval}(P_i, T_v) = 1$}
        \If{$\text{eval}(P_i, T_h) = 1$}
            \State \Return \textit{succeed} \Comment{Return ``succeed" if all tests pass}
        \Else
            \State \Return \textit{fail} \Comment{Return ``fail" if hidden tests fail}
        \EndIf
    \Else
        \State $feedback \gets \text{get\_feedback}(T_v, result)$ \Comment{Extract feedback from failed visible tests}
        \State $\text{expand}(node, feedback, node\_expansion)$ \Comment{Expand tree based on feedback}
        \State $\text{backpropagate}(result, node)$ \Comment{Backpropagate the result to update tree}
    \EndIf
    
    \State $i \gets i + 1$ \Comment{Increment iteration}
\EndWhile

\State \Return \textit{fail} \Comment{Return ``fail" if maximum iterations reached without passing all tests}
\end{algorithmic}
\end{algorithm}

\end{document}